\documentclass[runningheads]{llncs}

\usepackage{accv}

\usepackage{accvabbrv}

\usepackage{graphicx}
\usepackage{booktabs}

\usepackage{xcolor}
\usepackage{colortbl}
\usepackage{bbding}
\usepackage{subcaption} 

\usepackage[accsupp]{axessibility}  

\usepackage{hyperref}

\usepackage{orcidlink}

\begin{document}

\title{Hand-Object Contact Transition Detection in Egocentric Video}


\author{Huy Anh Nguyen \and
Chang Dong \and
Feras Dayoub \and Minh Hoai}

\authorrunning{H.A. Nguyen et al.}

\institute{Australian Institute for Machine Learning, Adelaide University, Adelaide, Australia}

\maketitle

\def\mA{\mathcal{A}}
\def\mB{\mathcal{B}}
\def\mC{\mathcal{C}}
\def\mD{\mathcal{D}}
\def\mE{\mathcal{E}}
\def\mF{\mathcal{F}}
\def\mG{\mathcal{G}}
\def\mH{\mathcal{H}}
\def\mI{\mathcal{I}}
\def\mJ{\mathcal{J}}
\def\mK{\mathcal{K}}
\def\mL{\mathcal{L}}
\def\mM{\mathcal{M}}
\def\mN{\mathcal{N}}
\def\mO{\mathcal{O}}
\def\mP{\mathcal{P}}
\def\mQ{\mathcal{Q}}
\def\mR{\mathcal{R}}
\def\mS{\mathcal{S}}
\def\mT{\mathcal{T}}
\def\mU{\mathcal{U}}
\def\mV{\mathcal{V}}
\def\mW{\mathcal{W}}
\def\mX{\mathcal{X}}
\def\mY{\mathcal{Y}}
\def\mZ{\mathcal{Z}}

\def\bbN{\mathbb{N}}
\def\bbR{\mathbb{R}}
\def\bbP{\mathbb{P}}
\def\bbQ{\mathbb{Q}}
\def\bbE{\mathbb{E}}

\def\1n{\mathbf{1}_n}
\def\0{\mathbf{0}}
\def\1{\mathbf{1}}

\def\A{{\bf A}}
\def\B{{\bf B}}
\def\C{{\bf C}}
\def\D{{\bf D}}
\def\E{{\bf E}}
\def\F{{\bf F}}
\def\G{{\bf G}}
\def\H{{\bf H}}
\def\I{{\bf I}}
\def\J{{\bf J}}
\def\K{{\bf K}}
\def\L{{\bf L}}
\def\M{{\bf M}}
\def\N{{\bf N}}
\def\O{{\bf O}}
\def\P{{\bf P}}
\def\Q{{\bf Q}}
\def\R{{\bf R}}
\def\S{{\bf S}}
\def\T{{\bf T}}
\def\U{{\bf U}}
\def\V{{\bf V}}
\def\W{{\bf W}}
\def\X{{\bf X}}
\def\Y{{\bf Y}}
\def\Z{{\bf Z}}

\def\a{{\bf a}}
\def\b{{\bf b}}
\def\c{{\bf c}}
\def\d{{\bf d}}
\def\e{{\bf e}}
\def\f{{\bf f}}
\def\g{{\bf g}}
\def\h{{\bf h}}
\def\i{{\bf i}}
\def\j{{\bf j}}
\def\k{{\bf k}}
\def\l{{\bf l}}
\def\m{{\bf m}}
\def\n{{\bf n}}
\def\o{{\bf o}}
\def\p{{\bf p}}
\def\q{{\bf q}}
\def\r{{\bf r}}
\def\s{{\bf s}}
\def\t{{\bf t}}
\def\u{{\bf u}}
\def\v{{\bf v}}
\def\w{{\bf w}}
\def\x{{\bf x}}
\def\y{{\bf y}}
\def\z{{\bf z}}

\def\balpha{\mbox{\boldmath{$\alpha$}}}
\def\bbeta{\mbox{\boldmath{$\beta$}}}
\def\bdelta{\mbox{\boldmath{$\delta$}}}
\def\bgamma{\mbox{\boldmath{$\gamma$}}}
\def\blambda{\mbox{\boldmath{$\lambda$}}}
\def\bsigma{\mbox{\boldmath{$\sigma$}}}
\def\btheta{\mbox{\boldmath{$\theta$}}}
\def\bomega{\mbox{\boldmath{$\omega$}}}
\def\bxi{\mbox{\boldmath{$\xi$}}}
\def\bnu{\mbox{\boldmath{$\nu$}}}
\def\bphi{\mbox{\boldmath{$\phi$}}}
\def\bmu{\mbox{\boldmath{$\mu$}}}

\def\bDelta{\mbox{\boldmath{$\Delta$}}}
\def\bOmega{\mbox{\boldmath{$\Omega$}}}
\def\bPhi{\mbox{\boldmath{$\Phi$}}}
\def\bLambda{\mbox{\boldmath{$\Lambda$}}}
\def\bSigma{\mbox{\boldmath{$\Sigma$}}}
\def\bGamma{\mbox{\boldmath{$\Gamma$}}}

\newcommand{\myprob}[1]{\mathop{\mathbb{P}}_{#1}}

\newcommand{\myexp}[1]{\mathop{\mathbb{E}}_{#1}}

\newcommand{\mydelta}[1]{1_{#1}}

\newcommand{\myminimum}[1]{\mathop{\textrm{minimum}}_{#1}}
\newcommand{\mymaximum}[1]{\mathop{\textrm{maximum}}_{#1}}
\newcommand{\mymin}[1]{\mathop{\textrm{minimize}}_{#1}}
\newcommand{\mymax}[1]{\mathop{\textrm{maximize}}_{#1}}
\newcommand{\mymins}[1]{\mathop{\textrm{min.}}_{#1}}
\newcommand{\mymaxs}[1]{\mathop{\textrm{max.}}_{#1}}
\newcommand{\myargmin}[1]{\mathop{\textrm{argmin}}_{#1}}
\newcommand{\myargmax}[1]{\mathop{\textrm{argmax}}_{#1}}
\newcommand{\myst}{\textrm{s.t. }}

\newcommand{\denselist}{\itemsep -1pt}
\newcommand{\sparselist}{\itemsep 1pt}

\definecolor{pink}{rgb}{0.9,0.5,0.5}
\definecolor{purple}{rgb}{0.5, 0.4, 0.8}
\definecolor{gray}{rgb}{0.3, 0.3, 0.3}
\definecolor{mygreen}{rgb}{0.2, 0.6, 0.2}

\newcommand{\cyan}[1]{\textcolor{cyan}{#1}}
\newcommand{\blue}[1]{\textcolor{blue}{#1}}
\newcommand{\magenta}[1]{\textcolor{magenta}{#1}}
\newcommand{\pink}[1]{\textcolor{pink}{#1}}
\newcommand{\green}[1]{\textcolor{green}{#1}}
\newcommand{\gray}[1]{\textcolor{gray}{#1}}
\newcommand{\mygreen}[1]{\textcolor{mygreen}{#1}}
\newcommand{\purple}[1]{\textcolor{purple}{#1}}

\definecolor{greena}{rgb}{0.4, 0.5, 0.1}
\newcommand{\greena}[1]{\textcolor{greena}{#1}}

\definecolor{bluea}{rgb}{0, 0.4, 0.6}
\newcommand{\bluea}[1]{\textcolor{bluea}{#1}}
\definecolor{reda}{rgb}{0.6, 0.2, 0.1}
\newcommand{\reda}[1]{\textcolor{reda}{#1}}

\def\changemargin#1#2{\list{}{\rightmargin#2\leftmargin#1}\item[]}
\let\endchangemargin=\endlist

\newcommand{\cm}[1]{}

\newcommand{\mhoai}[1]{{\color{blue}{[MH: #1]}}}
\newcommand{\ha}[1]{{\color{red}{[HA: #1]}}}
\newcommand{\rev}[1]{{\color{mygreen}{#1}}}

\newcommand{\tb}[1]{\textbf{#1}}
\newcommand{\ul}[1]{\underline{#1}}
\newcommand{\gain}[1]{\scriptsize\textcolor{blue}{(#1)}}
\newcommand{\lose}[1]{\scriptsize\textcolor{red}{(#1)}}

\newcommand{\mtodo}[1]{{\color{red}$\blacksquare$\textbf{[TODO: #1]}}}
\newcommand{\myheading}[1]{\vspace{1ex}\noindent \textbf{#1}}
\newcommand{\htimesw}[2]{\mbox{$#1$$\times$$#2$}}


\newif\ifshowsolution
\showsolutiontrue

\ifshowsolution
\newcommand{\Comment}[1]{\paragraph{\bf $\bigstar $ COMMENT:} {\sf #1} \bigskip}
\newcommand{\Solution}[2]{\paragraph{\bf $\bigstar $ SOLUTION:} {\sf #2} }
\newcommand{\Mistake}[2]{\paragraph{\bf $\blacksquare$ COMMON MISTAKE #1:} {\sf #2} \bigskip}
\else
\newcommand{\Solution}[2]{\vspace{#1}}
\fi

\newcommand{\truefalse}{
\begin{enumerate}
	\item True
	\item False
\end{enumerate}
}

\newcommand{\yesno}{
\begin{enumerate}
	\item Yes
	\item No
\end{enumerate}
}

\newcommand{\Sref}[1]{Sec.~\ref{#1}}
\newcommand{\Eref}[1]{Eq.~(\ref{#1})}
\newcommand{\Fref}[1]{Fig.~\ref{#1}}
\newcommand{\Tref}[1]{Table~\ref{#1}}

\newcommand{\parens}[1]{\left(#1\right)}
\newcommand{\braces}[1]{\left\{#1\right\}}
\newcommand{\bracks}[1]{\left[#1\right]}
\newcommand{\modulus}[1]{\left\vert#1\right\vert}
\newcommand{\norm}[1]{\left\Vert#1\right\Vert}
\newcommand{\angular}[1]{\langle#1\rangle}
\newcommand{\lmod}{\left|\!\left|}
\newcommand{\rmod}{\right|\!\right|}

\begin{abstract}
This paper tackles the task of hand-object contact transition detection in egocentric video, which aims to localize the two temporal boundaries of a contact episode: the moment a hand first establishes contact with an object (touch) and the moment contact is released (release). We propose a unified framework that jointly models the interacting hand and object to accurately localize both transition types. The proposed framework augments hand-centric spatiotemporal representations with object-aware features learned directly from the video and introduces a transition-aware event spotting formulation tailored to contact transitions, without relying on external object or hand-object interaction detectors during inference. To support this task, we present ContactTransition, a dataset comprising 4,021 egocentric videos with 8,498 touch and 8,077 release annotations, together with object-of-interest supervision. Experiments demonstrate that the proposed method consistently outperforms strong two-stage and end-to-end event-spotting baselines on both transition types. Finally, we show that detected contact transitions provide effective temporal anchors for downstream egocentric hand-object reasoning. Code is available at \url{https://github.com/bbvisual/oasis}.
\end{abstract}

\section{Introduction}
\label{sec:intro}

Hand-object manipulation is naturally organized into contact episodes that begin when the hand establishes contact with an object and end when the contact is released. These two transition moments define the temporal boundaries of every manipulation, enclosing the interval during which an object is grasped, transported, used, or repositioned. Detecting both transitions at frame-level precision provides a compact temporal representation of manipulation, enabling applications such as fine-grained action segmentation, grasp-duration analysis, skill assessment, and robot learning from demonstration. In this work, we study the task of detecting these two hand-object contact transitions in egocentric video. For brevity, we refer to the transition into contact as touch (contact onset) and the transition out of contact as release (contact offset).

Detecting contact transitions at frame-level precision is challenging because the visual evidence is often confined to only a few frames. Subtle changes in finger articulation, object motion, and the relative hand-object configuration must be distinguished from visually similar frames immediately before or after contact. Egocentric videos further complicate the problem through rapid camera motion, motion blur, frequent occlusions, and large variations in manipulation style. These challenges require jointly reasoning about fine-grained spatial cues and short-term temporal dynamics to accurately localize contact transitions.

Hand-object contact transition detection belongs to the broader class of precise temporal event detection, which has been widely studied in computer vision for tasks such as facial action unit onset detection and fine-grained action spotting. However, these methods are typically designed for settings that differ substantially from egocentric hand-object interaction. Facial event detection assumes static or near-static cameras, while sports action spotting operates on videos in which the target event occupies a large portion of the frame and the subject remains continuously visible. In contrast, hand-object contact transitions occur within compact regions around the interacting hand and object and depend on subtle changes in their spatial relationship.

Recent work\cite{hice} demonstrated that hand-centric spatiotemporal representations can effectively detect touch moments. However, it models only contact onset, leaving the complementary transition—contact release—unaddressed. Extending touch detection to contact transition detection is not simply a matter of adding another event class. Touch and release are intrinsically coupled, jointly defining the temporal boundaries of the same contact episode, yet they are characterized by fundamentally different visual evidence. Touch is naturally indicated by an approaching hand and the evolving grasp immediately before contact, making hand-centric motion cues highly informative. In contrast, after release, the hand often remains close to the object or even stays nearly stationary, while the released object may move, fall, or remain behind. Consequently, accurately localizing contact transitions requires reasoning about the evolving relationship between the hand and the object rather than the hand alone. Motivated by these observations, we propose a unified framework that jointly models both the interacting hand and object, while reformulating precise event spotting to explicitly account for the coupled nature of touch and release transitions.

To support research on hand-object contact transition detection, we introduce ContactTransition, a new dataset containing frame-accurate annotations of both touch and release transitions across 4,021 egocentric videos, comprising 8,498 touch and 8,077 release moments. Beyond benchmarking contact transition detection, the dataset enables studies of complete contact episodes and higher-level temporal reasoning over hand-object interactions. Experiments demonstrate that our framework consistently outperforms strong two-stage and end-to-end event-spotting baselines on both touch and release detection. Our contributions are summarized as follows:

\begin{enumerate}
\item We introduce the task of \emph{hand-object contact transition detection}, which aims to localize both touch (contact onset) and release (contact offset) moments in egocentric video.

\item We propose a unified framework for contact transition detection that models the interacting hand and object, together with a transition-aware event spotting formulation tailored to the coupled nature of touch and release.

\item We introduce \emph{ContactTransition}, a dataset containing 8,498 touch and 8,077 release annotations across 4,021 egocentric videos, enabling systematic evaluation of contact transition detection.

\item We demonstrate that contact transitions provide an effective temporal representation for downstream reasoning through \emph{HanDyVQA}, a recent benchmark for contact-aware video question answering.
\end{enumerate}

\section{Related Work}

\subsection{Hand-Object Interaction Understanding}

Understanding hands and their interactions with objects is fundamental to egocentric vision. Beyond a large body of work on detecting and tracking the hands themselves (e.g., \cite{m_Narasimhaswamy-etal-ICCV19, m_Huang-etal-CVPR22}), a long-standing goal is to understand how hands interact with objects \cite{fathiLearningRecognizeObjects2011,egoaction_survey}. A large body of image-based work studies contact as a per-frame attribute: 100DOH\cite{100doh}, ContactHands\cite{contacthands_2020}, and Hands23\cite{hands23} detect hands and classify their contact state from single frames, while BodyHands\cite{sn_bodyhands_cvpr_2022} infers hand--body association from spatial overlap. Beyond binary contact, HOT\cite{hot} localizes contact regions as dense heatmaps, and thermal- or pressure-based approaches\cite{contactdb,pressurevision,pressurevisionpp} recover fine-grained contact and force information from specialized supervision; contact maps have further been used to infer object functionality and affordances\cite{handprobe}. Video-based methods extend this analysis over time, detecting, segmenting, and tracking hands and interacted objects\cite{sn_hoist_cvpr_2024,ek_visor,hoi_detr,fuSequentialVotingRelational2022}, supported by richly annotated manipulation datasets\cite{hoi4d,taco,fanARCTICDatasetDexterous2023,hot3d}. Yagi et al. \cite{yagi2021hand} predict a binary contact state for each video frame from given hand-object tracks, but model contact as a state rather than localizing its transitions. Temporal structure, however, is typically addressed only at coarse granularity: large-scale egocentric benchmarks\cite{epickitchen,ego4d,egoexo4d} localize actions or object state changes at second-level tolerances, with Ego4D's point-of-no-return keyframe\cite{ego4d} being the closest to our setting, identifying a single state-change frame per clip rather than the paired boundaries of a contact episode. HiCE\cite{hice} recently showed that contact \emph{onset} can be localized with near frame-level precision, but the complementary transition out of contact remains unaddressed. Our work closes this gap by detecting both transitions, turning per-frame contact attributes into temporally delimited contact episodes.

Beyond perception, recent benchmarks probe whether video-language models genuinely understand hand-object interaction. HanDyVQA\cite{handyvqa} and related egocentric benchmarks\cite{egohoibench} evaluate fine-grained reasoning over manipulation dynamics, spatial relations, object parts, and state changes. Their findings show that current video and video-language models, including InternVideo2\cite{internvideo2}, LaViLa\cite{lavila}, and Qwen2.5-VL\cite{qwen25vl}, remain limited in contact-sensitive interaction understanding, motivating representations that better capture when and where hand-object evidence occurs.

\subsection{Precise Temporal Event Spotting}
Action spotting~\cite{soccernetas,soccernet} localizes the precise frame at which an event occurs, in contrast to temporal action localization\cite{sst,gtad,actionformer,tridet,asformer}, which predicts multi-frame intervals evaluated by temporal IoU. While interval-based methods can be adapted to short events, they are not well suited to instantaneous transitions that require frame-level precision\cite{as_survey}. Existing spotting methods generally follow two paradigms: two-stage approaches use pre-extracted features with temporal localization heads\cite{baidu,soares,astra,comedian,tsp,mstcn,asformer}, while end-to-end methods learn visual features jointly with the spotting objective\cite{e2e_spot,tdeed,uglf,astrm,adaspot}. Recent methods improve temporal precision through temporal shift modules, recurrent or encoder–decoder reasoning, and displacement refinement\cite{gsm,gsf,gru,tridet,tdeed}; HiCE\cite{hice} adapts this framework to hand contact by attending to hand regions.

However, most spotting formulations assume that events of different classes are temporally isolated, using mutually exclusive frame classification and a class-shared displacement for refinement. Contact transitions violate this assumption: touch and release bound the same episode, can occur close together, and may co-occur in bimanual manipulation. Our transition-aware formulation therefore scores each transition independently against a shared background reference and refines each transition with class-specific displacements.

\subsection{Object Grounding for Event Detection}
Several works incorporate object evidence into temporal localization. AOE-Net\cite{aam} models actor–object–environment interactions for temporal proposal generation, and UGLF\cite{uglf} grounds scene entities with vision-language detectors\cite{glip,glip2} to improve precise spotting. These approaches rely on external detectors at inference time, require a predefined object vocabulary, and localize objects unconditionally, irrespective of whether an object participates in the interaction. Detector-based pipelines for interacted objects\cite{sn_hoist_cvpr_2024,hoi_detr,fuSequentialVotingRelational2022} face similar constraints, and are least reliable precisely at contact moments, where the hand occludes the object it manipulates. In contrast, our framework learns to localize the object of interest from hand-conditioned features, using object annotations only as training-time supervision. At inference time, it does not rely on an object or hand-object interaction detector; the object regions are predicted within the contact-transition model and are therefore tied to the ongoing hand interaction.

\section{ContactTransition Dataset}
\label{sec:dataset}

Existing frame-precise contact annotations, such as TouchMoment \cite{hice}, focus on contact onset: the moment a hand first establishes contact with an object. While this enables precise touch detection, it does not capture the full temporal extent of a manipulation episode. To support hand-object contact transition detection, we introduce \emph{ContactTransition}, an egocentric video benchmark with frame-level annotations of both touch and release transitions. The benchmark follows the video sources, clip selection criteria, and touch annotation protocol used in prior frame-precise contact detection work \cite{hice}, while introducing release annotations that capture the endpoint of each contact episode.

\subsection{Release Annotation}

We define a release event as the first frame of sustained separation between the hand and the object involved in the interaction. Unlike touch, which is often indicated by the hand approaching and establishing contact with an object, release can be visually ambiguous. The hand may remain close to the object after letting go, the contact interface may be occluded by the hand, and short interruptions during re-grasping should not be treated as separate contact episodes. Therefore, annotators were instructed to use visible hand-object separation, object motion, and the surrounding temporal context to identify the earliest frame at which contact is no longer present.

To ensure temporal consistency, annotators considered the surrounding context when deciding whether a visible separation should be marked as a release. Brief separations caused by rapid hand motion, such as repeated tapping or quick adjustment of the same object, are treated as part of the same contact episode rather than as separate release–touch pairs. A release is marked only when the hand clearly disengages from the object, or when the subsequent contact corresponds to a distinct interaction. Ambiguous cases are excluded when the release moment cannot be localized reliably due to occlusion, motion blur, or insufficient visual evidence.

\begin{table}[t]
\centering
\footnotesize
\setlength{\tabcolsep}{5pt}
\begin{tabular}{@{}lrrrr@{}}
\toprule
      & \# Videos & \# Touch & \# Release & \# Events \\ \midrule
Train & 3,234      & 6,837     & 6,488       & 13,325     \\
Val   & 138       & 324      & 323        & 647       \\
Test  & 649       & 1,337     & 1,266       & 2,603      \\ \midrule
Total & 4,021      & 8,498     & 8,077       & 16,575     \\ \bottomrule
\end{tabular}
\vskip 0.05in
\caption{\textbf{ContactTransition statistics.} Touch and release counts are comparable overall. Release counts are slightly lower because some contact episodes continue beyond the clip boundary, and because two-hand interactions in TACO may contain separate contact onsets that terminate at a shared release moment.}
\label{table.data_statistics_v2}
\end{table}

\subsection{Expanded Manual Coverage}

ContactTransition uses manually verified annotations for all validation and test clips, including both HOI4D-derived clips and TACO segments. For the training split, prior frame-precise touch annotations provide reliable contact-onset labels for a large portion of the data \cite{hice}. However, we found that release timing is less amenable to the same automatic annotation strategy, since release events are more affected by occlusion, rapid re-contact, and subtle object motion. We therefore manually verify 35\% of the HOI4D training clips, prioritizing videos containing challenging release events. Across these and the validation/test clips, we annotated 5,126 release events and verified or corrected 1,668 touch events, excluding boundaries that could not be reliably localized.

\section{Proposed Methodology}
\label{sec:method}

In this section, we first formulate contact transition detection as an event-spotting task, where the goal is to localize the two boundaries of hand-object contact: touch onset and touch release. We then review the hand-centric representation used by HiCE~\cite{hice}, which provides the spatially localized hand context on which our model builds. Finally, we present \textbf{OASIS}, an object-aware spotting framework that augments hand-centric representations with candidate-object evidence and predicts touch and release through a transition-aware formulation.

\begin{figure}[!t]
    \centering
    \includegraphics[width=\linewidth]{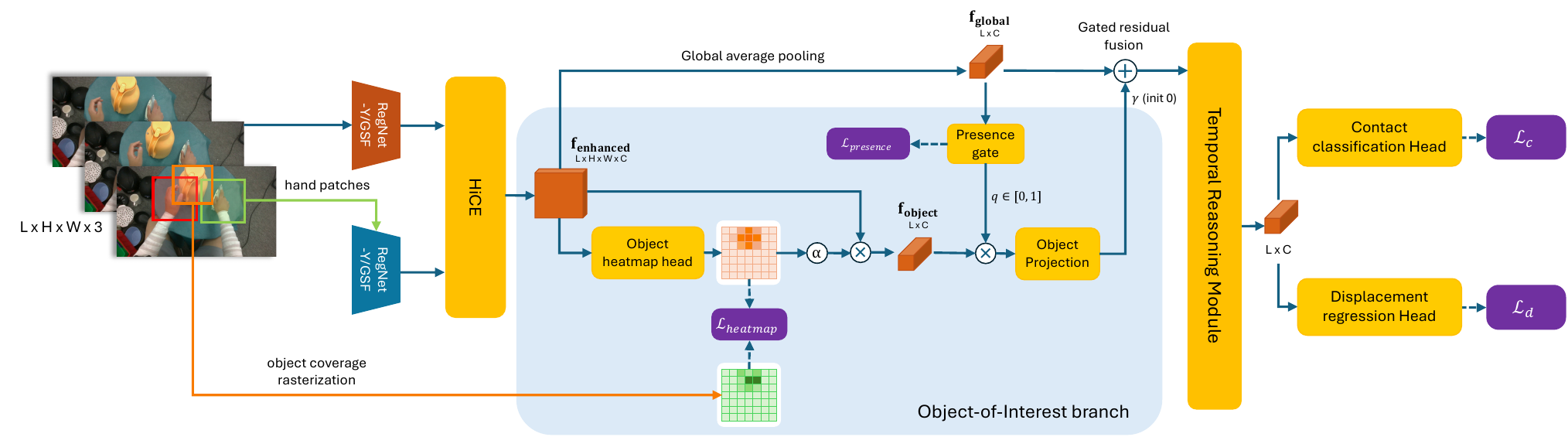}
    \caption{\textbf{Overview of OASIS.} Starting from the hand-enhanced feature map produced by HiCE \cite{hice}, OASIS predicts an object-of-interest heatmap and pools object-aware features from the predicted regions. A presence gate controls when object evidence should be used, and the resulting object-aware representation is fused into the temporal spotting stream. The final head predicts touch and release transitions with event scores and class-specific displacement refinement.}
    \label{fig.model_v2}
\end{figure}

\subsection{Problem Formulation}
\label{sec:problem}

Given an egocentric video clip of $L$ frames, $\mX=\{x_i\}_{i=1}^{L}$, the goal is to detect a sparse set of contact-transition events $\mathcal{E}=\{(t_k,c_k)\}_{k=1}^{K}$, where $t_k\in\{1,\ldots,L\}$ denotes the event frame and $c_k\in\mathcal{C}{=}\{\mathrm{touch},\mathrm{release}\}$ denotes the transition type.

Following the precise event-spotting protocol\cite{e2e_spot,hice}, a prediction is considered correct if it matches a ground-truth event of the same class within a temporal tolerance of $\delta$ frames. Since contact transitions are brief and visually subtle, we evaluate under strict tolerances, following the near-frame-level setting used in prior contact spotting work\cite{hice}.

Unlike touch-only detection, contact transition detection involves two coupled event types that define the same contact episode. However, they should not be forced into a mutually exclusive label space. In egocentric bimanual interaction, different hands may generate different transition events at the same frame, and these events may involve different objects. We therefore formulate touch and release as class-wise contact-transition events while using a shared object-aware representation to provide interaction context for both predictions.

\subsection{Preliminary: HiCE}
\label{sec:hice_prelim}

HiCE\cite{hice} is a hand-centric framework for precise contact onset detection. Given a video clip and hand boxes obtained from ground-truth annotations or an off-the-shelf hand detector such as Hands23, HiCE extracts two visual streams: a global frame feature that captures the surrounding scene, and localized hand-patch features that capture fine-grained hand appearance. The hand features are used to enhance the global frame representation through hand-to-frame attention, producing a hand-enhanced feature map $\mathcal{F}_{\mathrm{enhanced}}\in\mathbb{R}^{H\times W\times C}$. This representation preserves spatial information from the frame while emphasizing regions relevant to the observed hands.

HiCE then pools the enhanced feature map into a frame-level representation and feeds the resulting temporal sequence into a spotting head. The head predicts contact onset scores and temporal displacement offsets, which are used to refine event localization. This design is effective for touch detection because contact onset is often associated with visible hand motion and local hand-object configuration changes. However, it does not explicitly model the contacted object and is designed for a single transition type. In the following sections, we use $\mathcal{F}_{\mathrm{enhanced}}$ as the starting representation and introduce object-aware features and transition-aware prediction heads for detecting both touch and release.

\subsection{Object-of-Interest Branch}
\label{sec:object_branch}

The hand-enhanced feature map $\mathcal{F}_{\mathrm{enhanced}}$ captures fine-grained hand appearance and local hand context, but it does not explicitly identify the object involved in the interaction. This object information is important for contact transition detection. For touch, the contacted object helps disambiguate reaching from incidental hand motion; for release, it provides a spatial anchor when the hand remains close to the object or the change in hand pose is subtle. We therefore introduce an object-of-interest branch that localizes candidate interaction objects from $\mathcal{F}_{\mathrm{enhanced}}$ and injects object-aware evidence into the temporal spotting representation.

The branch predicts a class-agnostic object heatmap rather than object categories. This design reflects the goal of contact transition detection: the model only needs to know where the interaction-relevant object is, not what semantic class it belongs to. Since the heatmap is predicted from hand-enhanced features, object localization is conditioned on the observed hand context. The branch therefore learns to highlight objects likely to participate in the ongoing hand interaction, rather than performing generic object detection over the full scene.

We supervise the heatmap using object-of-interest annotations derived from available object annotations and contact-object annotations. For HOI4D, we use the provided object annotations; for TACO, we obtain per-frame contacted-object annotations using HOI-DETR\cite{hoi_detr}. Candidate-object regions are projected onto the hand-centered feature grid to form a soft coverage target $G\in[0,1]^{H\times W}$, where each cell indicates the degree of overlap with an interaction-relevant object. Given $\mathcal{F}_{\mathrm{enhanced}}\in\mathbb{R}^{H\times W\times C}$, a lightweight head predicts a heatmap $S=\phi_{\mathrm{object}}(\mathcal{F}_{\mathrm{enhanced}})\in\mathbb{R}^{H\times W}$ and is trained with a foreground-weighted binary cross-entropy:
\begin{equation}
    \mathcal{L}_{\mathrm{heatmap}} =
    \frac{1}{HW}\sum_j w_j\,\mathrm{BCE}(S_j, G_j).
\end{equation}
We set $w_j=1+(w_{\mathrm{fg}}-1)G_j$, with $w_{\mathrm{fg}}=10$, so empty cells receive weight 1 and fully covered cells receive weight 10. This compensates for the small foreground region relative to the background. To convert the heatmap into an object feature, we normalize the predicted object probabilities over the spatial grid and use them to pool from $\mathcal{F}_{\mathrm{enhanced}}$. This produces an object-aware feature $f_{\mathrm{object}}$ that summarizes the regions predicted to contain candidate interaction objects. We use normalized coverage weights rather than a softmax so that the pooling can cover the spatial extent of an object and can naturally support multiple candidate regions.

Because not every hand-centered crop contains a visible candidate object, we further predict an object-presence score $q$ from the frame representation using a two-layer MLP. The presence score gates the object feature before fusion, reducing the effect of noisy object readouts on frames where no reliable object evidence is visible. The gated object feature is injected into the global frame feature through a residual connection:
\begin{equation}
    f = f_{\mathrm{global}} + \gamma W_o(q f_{\mathrm{object}}),
\end{equation}
where $W_o$ is a learned projection and $\gamma$ is initialized to zero. The resulting feature sequence is passed to the temporal spotting module, allowing touch and release to be predicted from a shared hand- and object-aware representation.

\subsection{Transition-Aware Event Spotting}
\label{sec:transition_spotting}

Contact transition detection differs from standard action spotting because touch and release are coupled but not mutually exclusive. A softmax classifier forces the two transition types to compete for probability mass, which is undesirable when events occur close together or on the same frame under bimanual manipulation. We therefore predict each transition type independently against a shared background reference. Given logits $(z_l^{\mathrm{bg}}, z_l^{\mathrm{touch}}, z_l^{\mathrm{release}})$ at frame $l$, we define class-wise margins
\begin{equation}
    m_l^c = z_l^c - z_l^{\mathrm{bg}}, \qquad
    \hat{p}_l^c = \sigma(m_l^c),
    \qquad c\in\{\mathrm{touch},\mathrm{release}\}.
\end{equation}
Each transition type is trained with a focal binary cross-entropy against its own soft temporal target:
\begin{equation}
    \mathcal{L}_c =
    \frac{1}{LK}\sum_{l,c}
    \alpha_l^c(1-p_{t,l}^c)^{\gamma_f}
    \mathrm{BCE}(m_l^c,y_l^c),
\end{equation}
where $K=2$ and $y_l^c$ is the Gaussian target for class $c$. This formulation allows touch and release to be scored independently while sharing the same underlying representation.

We also make temporal displacement refinement class-aware. Prior spotting models regress a single displacement per frame, shared across all event classes. This is insufficient when touch and release are close in time, since one shared offset may refine one class toward the correct transition while moving the other away from it. Instead, the displacement head predicts $\hat{y}^{d}\in\mathbb{R}^{L\times K}$, with one offset per transition type. Each offset is supervised by the signed distance to the nearest event of the same class within radius $r$:
\begin{equation}
    y_l^{d,c} =
    \begin{cases}
    l-t_c^\ast(l), & |l-t_c^\ast(l)|\le r,\\
    0, & \text{otherwise},
    \end{cases}
\end{equation}
where $t_c^\ast(l)$ is the nearest class-$c$ event to frame $l$. The displacement loss is
\begin{equation}
    \mathcal{L}_d =
    \frac{1}{LK}\sum_{l,c}
    \big(\hat{y}_l^{d,c}-y_l^{d,c}\big)^2.
\end{equation}
At inference, each class score sequence is refined using its own displacement predictions, following the same temporal refinement procedure in \cite{hice}.

\subsection{Training Objective}
\label{sec:training}

The full model is trained with event classification, class-aware displacement regression, auxiliary grasp supervision inherited from HiCE, and object-of-interest supervision:
\begin{equation}
    \mathcal{L}
    =
    \mathcal{L}_c
    +
    \mathcal{L}_d
    +
    \lambda_g\mathcal{L}_g
    +
    \lambda_{\mathrm{object}}
    \left(
        \mathcal{L}_{\mathrm{heatmap}}
        +
        \lambda_p\mathcal{L}_{\mathrm{presence}}
    \right).
\end{equation}
Here, $\mathcal{L}_g$ regularizes the hand branch using grasp pseudo-labels, while $\mathcal{L}_{\mathrm{heatmap}}$ and $\mathcal{L}_{\mathrm{presence}}$ supervise candidate object localization and object presence estimation. The model is trained end-to-end; so hand-enhanced features, object localization, temporal reasoning, and touch-release prediction are optimized jointly.

\section{Experimental Analysis}
This section describes the experimental setup on ContactTransition, including implementation details, evaluation metrics, and the baseline methods used for comparison. We then present quantitative results followed by an ablation study to assess the contribution of each component in our approach.

\begin{table}[!h]
    \centering
    \setlength{\tabcolsep}{5pt}
    \footnotesize

    \subcaptionbox{Without non-maximum suppression (NMS).}{
    \resizebox{\linewidth}{!}{%
    \begin{tabular}{@{}l r|rrr|rrr@{}}
    \toprule
                &
                \multicolumn{1}{r}{}
                & \multicolumn{3}{c}{Touch}
                & \multicolumn{3}{c}{Release} \\
                \cmidrule(lr){3-5} \cmidrule(lr){6-8}
                & mAP
                & AP@0 & AP@1 & AP@2
                & AP@0 & AP@1 & AP@2 \\
    \midrule
    \multicolumn{8}{l}{\textit{(a) Two-stage methods}} \\
    TSP \cite{tsp} + MS-TCN \cite{mstcn}
                & 41.14 & 15.34 & 31.77 & 46.44 & 27.19 & \ul{57.82} & 68.26 \\
    TSP + ASFormer \cite{asformer}
                & 39.39 & \ul{17.16} & 30.65 & 40.40 & 26.32 & 54.41 & 67.40 \\
    V-JEPA \cite{vjepa2, vjepa2_1} + MS-TCN
                & \ul{43.93} & \tb{17.86} & \ul{41.20} & \ul{59.30} & 22.11 & 53.95 & 69.16 \\
    \midrule
    \multicolumn{8}{l}{\textit{(b) End-to-end methods}} \\
    E2ESpot \cite{e2e_spot}
                & 24.32 & 2.88  & 20.10 & 37.14 & 6.06  & 30.56 & 49.16 \\
    T-DEED \cite{tdeed}
                & 22.83 & 12.26 & 18.84 & 23.32 & 18.69 & 29.45  & 34.39 \\
    UGLF \cite{uglf}
                & 25.3 & 11.78 & 17.11 & 20.91 & 21.65 & 36.84 & 43.52 \\
    ASTRM \cite{astrm}
                & 21.48 & 13.56 & 19.09 & 22.22 & 18.51 & 25.96 & 29.54 \\
    AdaSpot \cite{adaspot}
                & 36.03 & 16.15 & 27.67 & 35.52 & \ul{27.81} & 50.36 & 58.66 \\
    HiCE \cite{hice}
                & 43.67 & 16.48 & 39.13 & 52.97 & 23.59 & 57.45 & \ul{72.41} \\
    \rowcolor{gray!20}
    OASIS (ours)
                & \tb{49.30} & 16.88 & \tb{44.86} & \tb{61.69} & \tb{29.42} & \tb{64.38} & \tb{78.54} \\
    \bottomrule
    \end{tabular}}}

    \vskip 0.1in

    \subcaptionbox{With NMS/Soft-NMS.}{
    \resizebox{\linewidth}{!}{%
    \begin{tabular}{@{}l r|rrr|rrr@{}}
    \toprule
                &
                \multicolumn{1}{r}{}
                & \multicolumn{3}{c}{Touch}
                & \multicolumn{3}{c}{Release} \\
                \cmidrule(lr){3-5} \cmidrule(lr){6-8}
                & mAP
                & AP@0 & AP@1 & AP@2
                & AP@0 & AP@1 & AP@2 \\
    \midrule
    \multicolumn{8}{l}{\textit{(a) Two-stage methods}} \\
    TSP \cite{tsp} + MS-TCN \cite{mstcn}
                & 49.19 & 9.96  & 41.82 & 66.93 & 20.77 & \ul{69.29} & \ul{86.37} \\
    TSP + ASFormer \cite{asformer}
                & \ul{49.28} & \ul{12.20} & 45.85 & 68.49 & 20.36 & 65.06 & 83.72 \\
    V-JEPA \cite{vjepa2, vjepa2_1} + MS-TCN
                & 47.94 & \tb{14.70} & \ul{46.40} & 69.42 & \ul{20.88} & 59.16 & 77.04 \\
    \midrule
    \multicolumn{8}{l}{\textit{(b) End-to-end methods}} \\
    E2ESpot \cite{e2e_spot}
                & 27.84 & 3.68  & 27.95 & 37.62 & 7.21  & 39.02 & 51.58 \\
    T-DEED \cite{tdeed}
                & 35.67 & 4.94  & 35.19 & 48.66 & 9.99  & 49.90 & 65.31 \\
    UGLF \cite{uglf}
                & 37.44 & 4.80  & 35.85 & 47.09 & 10.75 & 55.06 & 71.09 \\
    ASTRM \cite{astrm}
                & 38.92 & 6.61  & 36.11 & 60.03 & 11.68 & 47.83 & 71.23 \\
    AdaSpot \cite{adaspot}
                & 46.09 & 11.23 & 42.12 & 56.81 & \tb{21.56} & 65.77 & 79.05 \\
    HiCE \cite{hice}
                & 47.58 & 6.48 & 46.11 & \ul{71.00} & 11.32 & 64.79 & 85.76 \\
    \rowcolor{gray!20}
    OASIS (ours)
                & \tb{51.19} & 8.43 & \tb{51.41} & \tb{71.43} & 17.97 & \tb{70.90} & \tb{87.01} \\
    \bottomrule
    \end{tabular}}}

    \vskip 0.05in
    \caption{\textbf{Quantitative comparison on ContactTransition.}
Methods are grouped into two-stage approaches using pre-extracted features and end-to-end models. We report overall mAP, together with AP at frame tolerances $\delta \in \{0,1,2\}$ for touch and release detection. Bold and underline denote the best and second-best results.}
    \label{tab.results}
\end{table}

\subsection{Implementation Details}

We follow the training setup of HiCE~\cite{hice}. Models are trained on clips of $L=40$ frames for 50 epochs with AdamW~\cite{adamw}, using a batch size of 6, 5000 clips per epoch, an initial learning rate of $4{\times}10^{-4}$, a three-epoch warm-up, and cosine annealing. We use a RegNet-Y 800MF backbone~\cite{regnety} initialized from ImageNet-pretrained weights in \texttt{timm}~\cite{timm}. Hand patches are enlarged by a factor of 1.2 and resized to $224{\times}224$.

For event spotting, we use focal loss with $\alpha=0.9$ and $\gamma_f=2$, displacement supervision within $r=4$ frames, and a grasp loss weight of $\lambda_g=0.2$. The object-of-interest branch consists of a lightweight heatmap head and a presence gate, trained with loss weight $\lambda_{\mathrm{obj}}=0.15$ and presence weight $\lambda_p=0.75$. The presence threshold is set to $\tau=0.1$. Full architectural and optimization details are provided in the supplementary material.

\subsection{Evaluation Metrics}
We follow the precise event spotting protocol and evaluate touch and release as independent classes: for each class, a prediction is counted as correct if it falls within $\delta$ frames of an unmatched ground-truth event of that class, and we report per-class AP at tolerance thresholds $\delta \in \{0,1,2\}$ together with their mean (mAP). Frames may carry events of both classes, which per-class matching handles naturally. Unlike sports-based spotting benchmarks, where tolerances of \mbox{$\delta \leq 4$} are common, contact transitions in egocentric video occur in close temporal proximity, and release moments are particularly ambiguous at coarse tolerances due to gradual withdrawal; the low tolerance values emphasize near frame-level precision for both transition types.

\subsection{Baseline Methods}
We compare against baselines under two paradigms on ContactTransition, extending all methods from single-class touch spotting to the two-class touch--release setting by widening their classification heads accordingly. Two-stage approaches first extract features and then train a separate temporal model: we evaluate TSP\cite{tsp} paired with MS-TCN\cite{mstcn} and ASFormer\cite{asformer}, as well as V-JEPA\cite{vjepa2} paired with MS-TCN. TSP features are obtained by fine-tuning on our dataset following the original pretraining protocol, while V-JEPA features are extracted from the publicly available checkpoint. For end-to-end methods, we evaluate E2E-Spot\cite{e2e_spot}, T-DEED\cite{tdeed}, UGLF\cite{uglf}, ASTRM\cite{astrm}, AdaSpot\cite{adaspot}, and HiCE\cite{hice}, the strongest published method on the touch-only benchmark, extended to two classes with its original softmax formulation. Other action-spotting models such as ASTRA, COMEDIAN, and Soares\cite{astra,comedian,soares} require pre-extracted features from an ensemble of five networks\cite{baidu} trained on SoccerNet-v2 broadcast footage; these features are unavailable for egocentric video and retraining the ensemble is impractical, so a faithful reproduction of these methods on our benchmark is not possible.

For fair comparison, we apply Non-Maximum Suppression and Soft Non-Maximum Suppression (SNMS)\cite{softnms}, as used in T-DEED, to all methods with identical settings. Since suppression interacts differently with dense and sparse prediction styles, we additionally report all results without NMS/SNMS and discuss its effect on frame-exact precision in Sec.~\ref{sec:results}.

\subsection{Experiment Results}
\label{sec:results}
\cref{tab.results} presents contact transition detection performance on ContactTransition. Following the evaluation protocol of HiCE\cite{hice}, we report results both without and with NMS/SNMS to separate raw temporal precision from post-processed detection quality.

\myheading{Baseline comparison.}
Among two-stage methods, V-JEPA + MS-TCN achieves the highest mAP without NMS (43.93), while TSP + ASFormer leads with NMS (49.28), narrowly ahead of TSP + MS-TCN (49.19). Two-stage methods are strong at frame-exact touch localization, with V-JEPA + MS-TCN attaining the best touch AP@$\delta$=0 both without (17.86) and with NMS (14.70), reflecting the benefit of features trained or pretrained at clip level before temporal decoding. Among end-to-end baselines, HiCE is the strongest overall (43.67 without NMS, 47.58 with NMS), followed by AdaSpot (36.03 and 46.09). Most end-to-end methods gain substantially from suppression, with T-DEED improving from 22.83 to 35.67 and ASTRM from 21.48 to 38.92, indicating dense, redundant raw detections that NMS effectively prunes. Suppression consistently degrades frame-exact precision, however, and the effect is most severe on release: HiCE drops from 23.59 to 11.32 release AP@$\delta$=0, and AdaSpot, the best baseline at this operating point, falls from 27.81 to 21.56. This asymmetry is consistent with the gradual nature of withdrawal, which produces plateau-shaped score profiles around release moments that suppression merges or shifts.

\myheading{OASIS (Ours).}
Our method achieves the best overall mAP both without NMS (49.30) and with NMS (51.19), surpassing the strongest baseline in each setting by 5.37 and 1.91 mAP, respectively. Compared with HiCE, OASIS improves by 5.63 mAP without NMS and 3.61 mAP with NMS. The gains over HiCE concentrate on release detection, where the object-of-interest branch contributes +5.83 AP@$\delta$=0 (29.42 vs.\ 23.59), +6.93 AP@$\delta$=1, and +6.13 AP@$\delta$=2 without NMS, making OASIS the best method on release at every tolerance before suppression. This supports the motivation of Sec.\ \ref{sec:object_branch}: release is often signaled by subtle finger movements with little change in global hand pose, so an explicit object anchor provides evidence that hand-centric features lack. Touch localization also improves at relaxed tolerances, reaching 44.86 AP@$\delta$=1 and 61.69 AP@$\delta$=2 without NMS, the best among all methods, while frame-exact touch AP@$\delta$=0 (16.88) remains competitive with the two-stage leaders. Overall, OASIS provides the largest gains at $\delta \geq 1$ and on the release transition, precisely where hand cues are weakest and object grounding matters most.

\subsection{Ablation Study}

\begin{table}[t]
    \centering
    \setlength{\tabcolsep}{4pt}
    \footnotesize
    \resizebox{\linewidth}{!}{%
    \begin{tabular}{@{}l cc c cc c cc@{}}
    \toprule
    & \multicolumn{2}{c}{mAP} & & \multicolumn{2}{c}{Touch} & & \multicolumn{2}{c}{Release} \\
    \cmidrule(lr){2-3} \cmidrule(lr){5-6} \cmidrule(lr){8-9}
    & w/o NMS & w/ NMS & & AP@0 & AP@1 & & AP@0 & AP@1 \\
    \midrule
    OASIS (full)          & \tb{49.30} & \tb{51.19} & & \ul{16.88} & \tb{44.86} & & \tb{29.42} & \ul{64.38} \\
    \midrule
    w/o presence gate           & \ul{48.14} & 50.11 & & 16.80 & \ul{44.69} & & 27.69 & 63.59 \\
    softmax pooling (vs.\ CWP)  & 47.76 & \ul{50.37} & & \tb{17.78} & 42.77 & & \ul{29.41} & \tb{64.76} \\
    shared displacement (vs.\ per-class) & 44.61 & 47.52 & & 16.45 & 37.81 & & 25.87 & 59.78 \\
    w/o object branch           & 37.43 & 40.89 & & 13.68 & 29.08 & & 21.71 & 52.21 \\
    \midrule
    HiCE (softmax formulation)~\cite{hice} & 43.67 & 47.58 & & 16.48 & 39.13 & & 23.59 & 57.45 \\
    \bottomrule
    \end{tabular}}
    \vskip 0.05in
    \caption{Ablation study on ContactTransition. Each row removes or replaces one component of the full model. We report mean AP over both transitions and all tolerances $\delta \in \{0,1,2\}$, without and with NMS, together with per-transition AP at $\delta \in \{0,1\}$ without NMS. AP@2 columns are omitted for space and follow the same trends.}
    \label{tab.ablation}
\end{table}

\cref{tab.ablation} analyzes the contribution of each component. Removing the object branch causes the largest drop, reducing mAP with NMS from 51.19 to 40.89. This confirms that object grounding is central to contact transition detection, especially for release, where hand appearance alone is often ambiguous. The presence gate provides a smaller but consistent gain, improving mAP by 1.08 and increasing frame-exact release AP from 27.69 to 29.42, suggesting that suppressing unreliable object features is useful when no candidate object is visible in the hand-centered crop.

The pooling strategy also matters. Replacing coverage-weighted pooling (CWP) with softmax pooling reduces mAP, particularly before NMS, indicating that preserving spatial support over the predicted object region is preferable to concentrating on a single high-response cell. Class-aware displacement refinement is another important component: using a shared displacement reduces mAP by 3.67 and substantially lowers touch AP@1 from 44.86 to 37.81. This supports our argument that touch and release require separate temporal refinement because they can occur close together or even overlap in bimanual manipulation.

Finally, the original HiCE softmax formulation trails the full model by 3.61 mAP. Interestingly, removing the object branch from our transition-aware formulation performs worse than HiCE, showing that independent transition scoring alone is insufficient. Without object grounding, removing class competition can also remove useful suppression between ambiguous events. The full model benefits from the combination of independent transition scoring and object-aware evidence, indicating that the proposed components are complementary rather than merely additive.

\begin{figure}[!t]
    \centering
    \includegraphics[width=1\linewidth]{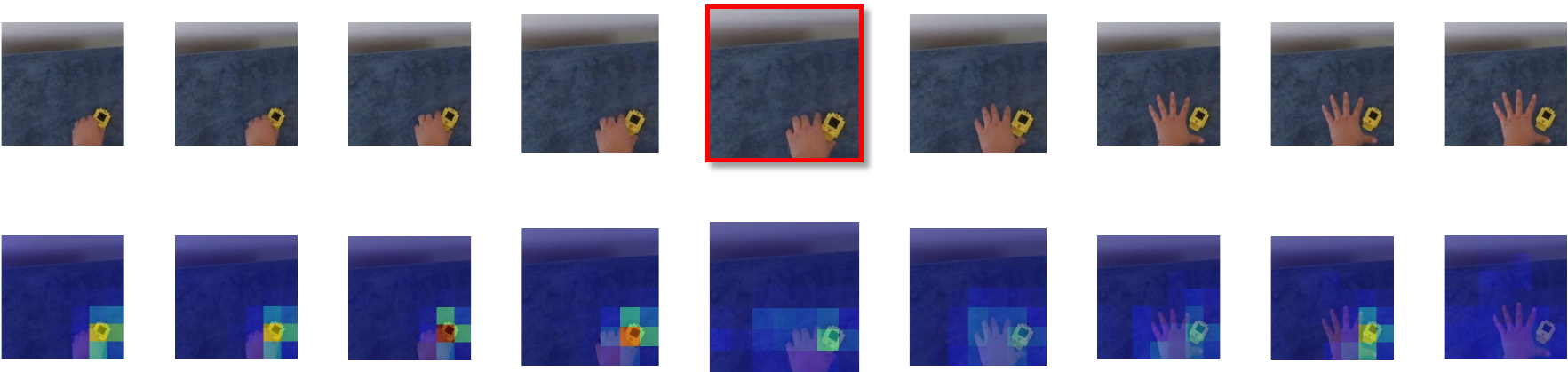}
    \caption{\textbf{Predicted object-of-interest heatmaps.}
Top: input frames around a contact transition, with the predicted transition frame highlighted in red. Bottom: heatmaps predicted by the object-of-interest branch of OASIS. Although object annotations are used only during training, the model learns to highlight the object involved in the hand interaction at inference time, providing object-aware evidence for precise touch--release localization.}
    \label{fig.qualitative}
\end{figure}

\section{Application: Contact Moments for Reasoning}
\label{sec:application}

Contact transitions can also serve as lightweight temporal cues for egocentric video reasoning. We evaluate this on HanDyVQA\cite{handyvqa}, a benchmark for fine-grained hand-object interaction understanding covering actions, object locations, manipulated objects, parts, processes, and state changes. We use Qwen2.5-VL\cite{qwen25vl} with uniform frame sampling as the base model, and compare it with a contact-aware variant that additionally provides predicted touch and release scores as temporal context together with the question.

As shown in \cref{tab:handyvqa_application}, adding contact context improves the average score from 58.01 to 58.49 without modifying the VLM or changing the sampled frames. The largest gains appear on part-level and state reasoning, suggesting that contact-transition scores help the model identify when fine-grained hand-object evidence is likely to be relevant. The process category decreases slightly, indicating that transition timing alone does not capture all procedural cues.

\begin{table}[t]
    \centering
    \footnotesize
    \setlength{\tabcolsep}{4pt}
    \resizebox{\linewidth}{!}{%
    \begin{tabular}{@{}lccccccc@{}}
    \toprule
    Method & Action & Location & Objects (AP) & Parts & Process & State & Avg \\
    \midrule
    Qwen2.5-VL & 58.66 & 53.12 & 72.28 & 50.36 & 54.88 & 58.75 & 58.01 \\
    Qwen2.5-VL + contact context & 58.84 & 53.94 & 72.63 & 51.59 & 54.06 & 59.89 & 58.49 \\
    \midrule
    $\Delta$ (improvement) & +0.18 & +0.82 & +0.35 & +1.23 & -0.82 & +1.14 & +0.48 \\
    \bottomrule
    \end{tabular}}
    \vskip 0.05in
    \caption{\textbf{Contact transitions as temporal context for HanDyVQA.} Both settings use Qwen2.5-VL with uniform frame sampling; the second row additionally provides predicted touch and release scores as temporal context.}
    \label{tab:handyvqa_application}
    \vspace{-0.12in}
\end{table}

\section{Conclusions}
\label{sec:conclusion}

We introduced hand–object contact transition detection, the task of precisely localizing the touch and release boundaries of contact episodes in egocentric video. To address this task, we proposed OASIS, which combines hand- and object-aware representations with transition-aware spotting to jointly detect both boundaries, without requiring external object or hand–object interaction detectors at inference.

We also introduced ContactTransition, a benchmark providing frame-level touch and release annotations together with object-of-interest supervision. OASIS consistently outperforms strong two-stage and end-to-end spotting baselines, while our downstream experiments demonstrate that detected contact transitions provide effective temporal anchors for egocentric hand–object reasoning. Extending the framework to per-hand transition attribution, leveraging annotations that identify the hand involved in each contact episode, is a promising direction for future work. Overall, our results establish contact transitions as a compact and useful representation for fine-grained manipulation understanding.

{\small    \myheading{Acknowledgements:}} This work was funded by the Australian Institute for Machine Learning (Adelaide University) and the Centre for Augmented Reasoning, an initiative by the Department of Education, Australian Government.
%
%
\bibliographystyle{splncs04}
\bibliography{longstrings, hand_touch}

@inproceedings{m_Narasimhaswamy-etal-ICCV19, 
  author = {Supreeth Narasimhaswamy and Zhengwei Wei and Yang Wang and Justin Zhang and Minh Hoai}, 
  title = {Contextual Attention for Hand Detection in the Wild}, 
  year = {2019}, 
  booktitle = iccv, 
}

@inproceedings{m_Huang-etal-CVPR22, 
  author = {Mingzhen Huang and Supreeth Narasimhaswamy and Saif Vazir and  Haibin Ling and Minh Hoai}, 
  title = {Forward Propagation, Backward Regression, and Pose Association for Hand Tracking in the Wild}, 
  year = {2022}, 
  booktitle = cvpr, 
}

@inproceedings{fanARCTICDatasetDexterous2023,
  author      = {Fan, Zicong and Taheri, Omid and Tzionas, Dimitrios and Kocabas, Muhammed and Kaufmann, Manuel and Black, Michael J. and Hilliges, Otmar},
  title       = {{{ARCTIC}}: {{A Dataset}} for {{Dexterous Bimanual Hand-Object Manipulation}}},
  booktitle   = cvpr,
  year        = {2023},
}

@inproceedings{mstcn,
  author      = {Farha, Yazan Abu and Gall, Jürgen},
  title       = {MS-TCN: Multi-Stage Temporal Convolutional Network for Action Segmentation},
  booktitle   = cvpr,
  year        = {2019},
}

@inproceedings{fathiLearningRecognizeObjects2011,
  author      = {Fathi, Alireza and Ren, Xiaofeng and Rehg, James M.},
  title       = {Learning to Recognize Objects in Egocentric Activities},
  booktitle   = cvpr,
  year        = {2011},
}

@inproceedings{fuSequentialVotingRelational2022,
  author      = {Fu, Qichen and Liu, Xingyu and Kitani, Kris},
  title       = {Sequential Voting With Relational Box Fields for Active Object Detection},
  booktitle   = cvpr,
  year        = {2022},
}

@inproceedings{e2e_spot,
  author      = {Hong, James and Zhang, Haotian and Gharbi, Micha{\"e}l and Fisher, Matthew and Fatahalian, Kayvon},
  title       = {Spotting {{Temporally Precise}}, {{Fine-Grained Events}} in {{Video}}},
  booktitle   = eccv,
  year        = {2022},
}

@inproceedings{taco,
  author      = {Liu, Yun and Yang, Haolin and Si, Xu and Liu, Ling and Li, Zipeng and Zhang, Yuxiang and Liu, Yebin and Yi, Li},
  title       = {{{TACO}}: Benchmarking Generalizable Bimanual Tool-ACtion-Object Understanding},
  booktitle   = cvpr,
  year        = {2024},
}

@inproceedings{uglf,
  author      = {Tran, Kim Hoang and Do, Phuc Vuong and Ly, Ngoc Quoc and Le, Ngan},
  title       = {Unifying {{Global}} and {{Local Scene Entities Modelling}} for {{Precise Action Spotting}}},
  booktitle   = ijcnn,
  year        = {2024},
}

@inproceedings{tdeed,
  author      = {Xarles, Artur and Escalera, Sergio and Moeslund, Thomas B. and Clap{\'e}s, Albert},
  title       = {T-{{DEED}}: {{Temporal-Discriminability Enhancer Encoder-Decoder}} for {{Precise Event Spotting}} in {{Sports Videos}}},
  booktitle   = cvprw,
  year        = {2024},
}

@inproceedings{regnety,
  author      = {Radosavovic, Ilija and Kosaraju, Raj Prateek and Girshick, Ross and He, Kaiming and Dollár, Piotr},
  title       = {Designing Network Design Spaces},
  booktitle   = cvpr,
  year        = {2020},
}

@inproceedings {gsm,
  author      = { Sudhakaran, Swathikiran and Escalera, Sergio and Lanz, Oswald },
  title       = {{ Gate-Shift Networks for Video Action Recognition }},
  booktitle   = cvpr,
  year        = {2020}
}

@inproceedings{hands23,
  author      = {Cheng, Tianyi and Shan, Dandan and Hassen, Ayda and Higgins, Richard E. L. and Fouhey, David F.},
  title       = {Towards a richer 2D understanding of hands at scale},
  booktitle   = nips,
  year        = {2023},
}

@inproceedings{contacthands_2020,
  author      = {Supreeth Narasimhaswamy and Trung Nguyen and Minh Hoai},
  title       = {Detecting Hands and Recognizing Physical Contact in the Wild},
  booktitle   = nips,
  year        = {2020},
}

@inproceedings {100doh,
  author      = { Shan, Dandan and Geng, Jiaqi and Shu, Michelle and Fouhey, David F. },
  title       = {{ Understanding Human Hands in Contact at Internet Scale }},
  booktitle   = cvpr,
  year        = {2020},
}

@inproceedings{tridet,
  author      = {Shi, Dingfeng and Zhong, Yujie and Cao, Qiong and Ma, Lin and Li, Jia and Tao, Dacheng},
  title       = {{{TriDet}}: Temporal Action Detection with Relative Boundary Modeling},
  booktitle   = cvpr,
  year        = {2023},
}

@inproceedings{astra,
  author      = {Xarles, Artur and Escalera, Sergio and Moeslund, Thomas B. and Clap{\'e}s, Albert},
  title       = {ASTRA: An Action Spotting TRAnsformer for Soccer Videos},
  booktitle   = {Proceedings of the 6th International Workshop on Multimedia Content Analysis in Sports},
  year        = {2023}
}

@inproceedings{comedian,
  author      = {Denize, Julien and Liashuha, Mykola and Rabarisoa, Jaonary and Orcesi, Astrid and H\'erault, Romain},
  title       = {COMEDIAN: Self-Supervised Learning and Knowledge Distillation for Action Spotting Using Transformers},
  booktitle   = wacvw,
  year        = {2024}
}

@inproceedings{adamw,
  author      = {Ilya Loshchilov and Frank Hutter},
  title       = {Decoupled Weight Decay Regularization},
  booktitle   = iclr,
  year        = {2019},
}

@inproceedings{softnms,
  title       = {Soft-NMS — Improving Object Detection with One Line of Code},
  author      = {Navaneeth Bodla and Bharat Singh and Rama Chellappa and Larry S. Davis},
  booktitle   = iccv,
  year        = {2017},
}

@inproceedings {contactdb,
  author      = { Brahmbhatt, Samarth and Ham, Cusuh and Kemp, Charles C. and Hays, James },
  booktitle   = cvpr,
  title       = {{ ContactDB: Analyzing and Predicting Grasp Contact via Thermal Imaging }},
  year        = {2019},
}

@inproceedings{pressurevision,
  author      = {Grady, Patrick and Tang, Chengcheng and Brahmbhatt, Samarth and Twigg, Christopher D. and Wan, Chengde and Hays, James and Kemp, Charles C.},
  title       = {PressureVision: Estimating Hand Pressure from~a~Single RGB Image},
  booktitle   = eccv,
  year        = {2022},
}

@inproceedings{pressurevisionpp,
  author      = {Grady, Patrick and Collins, Jeremy A and Tang, Chengcheng and Twigg, Christopher D and Aneja, Kunal and Hays, James and Kemp, Charles C},
  title       = {{PressureVision++}: Estimating Fingertip Pressure from Diverse RGB Images},
  booktitle   = wacv,
  year        = {2024},
}

@inproceedings{hoi4d,
  author      = {Liu, Yunze and Liu, Yun and Jiang, Che and Lyu, Kangbo and Wan, Weikang and Shen, Hao and Liang, Boqiang and Fu, Zhoujie and Wang, He and Yi, Li},
  title       = {{{HOI4D}}: A 4D Egocentric Dataset for Category-Level Human-Object Interaction},
  booktitle   = cvpr,
  year        = {2022},
}

@inproceedings{hot3d,
  author      = {Banerjee, Prithviraj and Shkodrani, Sindi and Moulon, Pierre and Hampali, Shreyas and Han, Shangchen and Zhang, Fan and Zhang, Linguang and Fountain, Jade and Miller, Edward and Basol, Selen and Newcombe, Richard and Wang, Robert and Engel, Jakob Julian and Hodan, Tomas},
  title       = {HOT3D: Hand and Object Tracking in 3D from Egocentric Multi-View Videos},
  booktitle   = cvpr,
  year        = {2025},
  }

@inproceedings{ego4d,
  author      ={Grauman, Kristen and Westbury, Andrew and Byrne, Eugene and Chavis, Zachary and Furnari, Antonino and Girdhar, Rohit and Hamburger, Jackson and Jiang, Hao and Liu, Miao and Liu, Xingyu and Martin, Miguel and Nagarajan, Tushar and Radosavovic, Ilija and Ramakrishnan, Santhosh Kumar and Ryan, Fiona and Sharma, Jayant and Wray, Michael and Xu, Mengmeng and Xu, Eric Zhongcong and Zhao, Chen and Bansal, Siddhant and Batra, Dhruv and Cartillier, Vincent and Crane, Sean and Do, Tien and Doulaty, Morrie and Erapalli, Akshay and Feichtenhofer, Christoph and Fragomeni, Adriano and Fu, Qichen and Gebreselasie, Abrham and González, Cristina and Hillis, James and Huang, Xuhua and Huang, Yifei and Jia, Wenqi and Khoo, Weslie and Koláĭ, Jáchym and Kottur, Satwik and Kumar, Anurag and Landini, Federico and Li, Chao and Li, Yanghao and Li, Zhenqiang and Mangalam, Karttikeya and Modhugu, Raghava and Munro, Jonathan and Murrell, Tullie and Nishiyasu, Takumi and Price, Will and Puentes, Paola Ruiz and Ramazanova, Merey and Sari, Leda and Somasundaram, Kiran and Southerland, Audrey and Sugano, Yusuke and Tao, Ruijie and Vo, Minh and Wang, Yuchen and Wu, Xindi and Yagi, Takuma and Zhao, Ziwei and Zhu, Yunyi and Arbeláez, Pablo and Crandall, David and Damen, Dima and Farinella, Giovanni Maria and Fuegen, Christian and Ghanem, Bernard and Ithapu, Vamsi Krishna and Jawahar, C. V. and Joo, Hanbyul and Kitani, Kris and Li, Haizhou and Newcombe, Richard and Oliva, Aude and Park, Hyun Soo and Rehg, James M. and Sato, Yoichi and Shi, Jianbo and Shou, Mike Zheng and Torralba, Antonio and Torresani, Lorenzo and Yan, Mingfei and Malik, Jitendra},
  title       = {Ego4D: Around the World in 3,000 Hours of Egocentric Video},
  booktitle   = cvpr,
  year        = {2022},
}

@inproceedings{egoexo4d,
  author      = {Grauman, Kristen and Westbury, Andrew and Torresani, Lorenzo and Kitani, Kris and Malik, Jitendra and Afouras, Triantafyllos and Ashutosh, Kumar and Baiyya, Vijay and Bansal, Siddhant and Boote, Bikram and Byrne, Eugene and Chavis, Zach and Chen, Joya and Cheng, Feng and Chu, Fu-Jen and Crane, Sean and Dasgupta, Avijit and Dong, Jing and Escobar, Maria and Forigua, Cristhian and Gebreselasie, Abrham and Haresh, Sanjay and Huang, Jing and Islam, Md Mohaiminul and Jain, Suyog and Khirodkar, Rawal and Kukreja, Devansh and Liang, Kevin J and Liu, Jia-Wei and Majumder, Sagnik and Mao, Yongsen and Martin, Miguel and Mavroudi, Effrosyni and Nagarajan, Tushar and Ragusa, Francesco and Ramakrishnan, Santhosh Kumar and Seminara, Luigi and Somayazulu, Arjun and Song, Yale and Su, Shan and Xue, Zihui and Zhang, Edward and Zhang, Jinxu and Castillo, Angela and Chen, Changan and Fu, Xinzhu and Furuta, Ryosuke and Gonzalez, Cristina and Gupta, Prince and Hu, Jiabo and Huang, Yifei and Huang, Yiming and Khoo, Weslie and Kumar, Anush and Kuo, Robert and Lakhavani, Sach and Liu, Miao and Luo, Mi and Luo, Zhengyi and Meredith, Brighid and Miller, Austin and Oguntola, Oluwatumininu and Pan, Xiaqing and Peng, Penny and Pramanick, Shraman and Ramazanova, Merey and Ryan, Fiona and Shan, Wei and Somasundaram, Kiran and Song, Chenan and Southerland, Audrey and Tateno, Masatoshi and Wang, Huiyu and Wang, Yuchen and Yagi, Takuma and Yan, Mingfei and Yang, Xitong and Yu, Zecheng and Zha, Shengxin Cindy and Zhao, Chen and Zhao, Ziwei and Zhu, Zhifan and Zhuo, Jeff and Arbelaez, Pablo and Bertasius, Gedas and Damen, Dima and Engel, Jakob and Farinella, Giovanni Maria and Furnari, Antonino and Ghanem, Bernard and Hoffman, Judy and Jawahar, C.V. and Newcombe, Richard and Park, Hyun Soo and Rehg, James M. and Sato, Yoichi and Savva, Manolis and Shi, Jianbo and Shou, Mike Zheng and Wray, Michael},
  title       = {Ego-Exo4D: Understanding Skilled Human Activity from First- and Third-Person Perspectives},
  booktitle   = cvpr,
  year        = {2024},
}

@inproceedings{glip,
  author      = {Liunian Harold Li and Pengchuan Zhang and Haotian Zhang and Jianwei Yang and Chunyuan Li and Yiwu Zhong and Lijuan Wang and Lu Yuan and Lei Zhang and Jenq-Neng Hwang and Kai-Wei Chang and Jianfeng Gao},
  title       = {Grounded Language-Image Pre-training},
  booktitle   = cvpr,
  year        = {2022},
}

@inproceedings{glip2,
  author      = {Zhang, Haotian and Zhang, Pengchuan and Hu, Xiaowei and Chen, Yen-Chun and Li, Liunian and Dai, Xiyang and Wang, Lijuan and Yuan, Lu and Hwang, Jenq-Neng and Gao, Jianfeng},
  title       = {GLIPv2: Unifying Localization and Vision-Language Understanding},
  booktitle   = nips,
  year        = {2022},
}

@inproceedings{soccernet,
  author      = {Deliège, Adrien and Cioppa, Anthony and Giancola, Silvio and Seikavandi, Meisam J. and Dueholm, Jacob V. and Nasrollahi, Kamal and Ghanem, Bernard and Moeslund, Thomas B. and Van Droogenbroeck, Marc},
  title       = {SoccerNet-v2: A Dataset and Benchmarks for Holistic Understanding of Broadcast Soccer Videos},
  booktitle   = cvprw,
  year        = {2021},
}

@inproceedings{sn_hoist_cvpr_2024,
	author      = {Supreeth Narasimhaswamy and Huy Nguyen and Lihan Huang and Minh Hoai},
	title       = {HOIST-Former: Hand-held Objects Identification, Segmentation, and Tracking in the Wild},
	booktitle   = cvpr,
	year        = {2024},
}

@inproceedings{sn_bodyhands_cvpr_2022,
	author      = {Supreeth Narasimhaswamy and Thanh Nguyen and Mingzhen Huang and Minh Hoai},
	title       = {Whose Hands Are These? Hand Detection and Hand-Body Association in the Wild},
	booktitle   = cvpr,
	year        = {2022},
}

@inproceedings{gtad,
  author      = {Xu, Mengmeng and Zhao, Chen and Rojas, David S. and Thabet, Ali and Ghanem, Bernard},
  title       = {{G-TAD}: Sub-Graph Localization for Temporal Action Detection},
  booktitle   = cvpr,
  year        = {2020},
}

@inproceedings{asformer,
  title       = {ASFormer: Transformer for Action Segmentation},
  author      = {Fangqiu Yi and Hongyu Wen and Tingting Jiang},
  booktitle   = bmvc,
  year        = {2021},
}

@inproceedings{sst,
  author      = {Buch, Shyamal and Escorcia, Victor and Shen, Chuanqi and Ghanem, Bernard and Carlos Niebles, Juan},
  title       = {{SST}: Single-Stream Temporal Action Proposals},
  booktitle   = cvpr,
  year        = {2017},
}

@inproceedings{actionformer,
  title       = {ActionFormer: Localizing Moments of Actions with Transformers},
  author      = {Zhang, Chen-Lin and Wu, Jianxin and Li, Yin},
  booktitle   = eccv,
  year        = {2022}
}

@inproceedings{soccernetas,
  title       = {SoccerNet: A Scalable Dataset for Action Spotting in Soccer Videos},
  author      = {Silvio Giancola and Mohieddine Amine and Tarek Dghaily and Bernard Ghanem},
  booktitle   = cvprw,
  year        = {2018},
}

@inproceedings{hot,
  author      = {Chen, Yixin and Dwivedi, Sai Kumar and Black, Michael J. and Tzionas, Dimitrios},
  title       = {Detecting Human-Object Contact in Images},
  booktitle   = cvpr,
  year        = {2023},
}

@inproceedings{handprobe,
  author      = {Mohit Goyal and Sahil Modi and Rishabh Goyal and Saurabh Gupta},
  title       = {Human Hands as Probes for Interactive Object Understanding},
  booktitle   = cvpr,
  year        = {2022},
}

@inproceedings{ek_visor,
  author      = {Darkhalil, Ahmad and Shan, Dandan and Zhu, Bin and Ma, Jian and Kar, Amlan and Higgins, Richard and Fidler, Sanja and Fouhey, David and Damen, Dima},
  title       = {EPIC-KITCHENS VISOR Benchmark: VIdeo Segmentations and Object Relations},
  booktitle   = {Advances in Neural Information Processing Systems Track on Datasets and Benchmarks},
  year        = {2022}
}

@inproceedings{tsp,
  author      = {Alwassel, Humam and Giancola, Silvio and Ghanem, Bernard},
  title       = {TSP: Temporally-Sensitive Pretraining of Video Encoders for Localization Tasks},
  booktitle   = iccvw,
  year        = {2021}
}

@inproceedings{astrm,
  author      = {Santra, Sanchayan and Chudasama, Vishal and Wasnik, Pankaj and Balasubramanian, Vineeth N},
  title       = {Precise Event Spotting in Sports Videos: Solving Long-Range Dependency and Class Imbalance},
  booktitle   = cvpr,
  year        = {2025},
}

@inproceedings{adaspot,
  author      = {Xarles, Artur and Escalera, Sergio and Moeslund, Thomas B and Clap{\'e}s, Albert},
  title       = {AdaSpot: Spend Resolution Where It Matters for Precise Event Spotting},
  booktitle   = cvpr,
  year        = {2026},
}

@inproceedings{hice,
  author      = {Huy Anh Nguyen and Feras Dayoub and Minh Hoai},
  title       = {Detecting Precise Hand Touch Moments in Egocentric Video},
  booktitle   = cvprf,
  year        = {2026},
}

@inproceedings{handyvqa,
  author      = {Tateno, Masatoshi and Kato, Gido and Kataoka, Hirokatsu and Sato, Yoichi and Yagi, Takuma},
  title       = {HanDyVQA: A Video QA Benchmark for Fine-Grained Hand-Object Interaction Dynamics},
  booktitle   = cvpr,
  year        = {2026},
}

@inproceedings{egohoibench,
  author      = {Boshen Xu and Ziheng Wang and Yang Du and Zhinan Song and Sipeng Zheng and Qin Jin},
  title       = {Do Egocentric Video-Language Models Truly Understand Hand-Object Interactions?},
  booktitle   = iclr,
  year        = {2025}
}

@inproceedings{internvideo2,
  author      = {Wang, Yi and Li, Kunchang and Li, Xinhao and Yu, Jiashuo and He, Yinan and Chen, Guo and Pei, Baoqi and Zheng, Rongkun and Wang, Zun and Shi, Yansong and Jiang, Tianxiang and Li, Songze and Xu, Jilan and Zhang, Hongjie and Huang, Yifei and Qiao, Yu and Wang, Yali and Wang, Limin},
  title       = {InternVideo2: Scaling Foundation Models for Multimodal Video Understanding},
  booktitle   = eccv,
  year        = {2024},
}

@inproceedings{lavila,
  author      = {Zhao, Yue and Misra, Ishan and Kr\"ahenb\"uhl, Philipp and Girdhar, Rohit},
  title       = {Learning Video Representations From Large Language Models},
  booktitle   = cvpr,
  year        = {2023},
}

@inproceedings{yagi2021hand,
  title       = {Hand-Object Contact Prediction via Motion-Based Pseudo-Labeling and Guided Progressive Label Correction},
  author      = {Yagi, Takuma and Hasan, Md. Tasnimul and Sato, Yoichi},
  booktitle   = bmvc,
  year        = {2021},
}

@article{gsf,
  author      ={Sudhakaran, Swathikiran and Escalera, Sergio and Lanz, Oswald},
  title       ={Gate-Shift-Fuse for Video Action Recognition},
  journal     = tpami,
  volume      = {45},
  number      = {9},
  pages       = {10913--10928},
  year        = {2023},
}

@article{epickitchen,
  author      = {Damen, Dima and Doughty, Hazel and Farinella, Giovanni Maria and Furnari, Antonino and Kazakos, Evangelos and Ma, Jian and Moltisanti, Davide and Munro, Jonathan and Perrett, Toby and Price, Will and Wray, Michael},
  title       = {Rescaling Egocentric Vision: Collection, Pipeline and Challenges for EPIC-KITCHENS-100},
  journal     = ijcv,
  volume      = {130},
  number      = {1},
  pages       = {33--55},
  year        = {2022},
}

@article{egoaction_survey,
  author      = {Adrián Núñez-Marcos and Gorka Azkune and Ignacio Arganda-Carreras},
  title       = {Egocentric Vision-based Action Recognition: A survey},
  journal     = {Neurocomputing},
  volume      = {472},
  pages       = {175--197},
  year        = {2022},
}

@article{aam,
  author      = {Vo, Khoa and Truong, Sang and Yamazaki, Kashu and Raj, Bhiksha and Tran, Minh-Triet and Le, Ngan},
  title       = {AOE-Net: Entities Interactions Modeling with Adaptive Attention Mechanism for Temporal Action Proposals Generation},
  journal     = ijcv,
  volume      = {131},
  number      = {1},
  pages       = {302--323},
  year        = {2023},
}

@misc{as_survey,
  author        = {Xu, Hao and Baniya, Arbind Agrahari and Well, Sam and Bouadjenek, Mohamed Reda and Dazeley, Richard and Aryal, Sunil},
  title         = {Deep Learning for Sports Video Event Detection: Tasks, Datasets, Methods, and Challenges},
  howpublished  = {arXiv preprint arXiv:2505.03991},
  year          = {2025},
}

@misc{soares,
  author        = {João V. B. Soares and Avijit Shah},
  title         = {Action Spotting using Dense Detection Anchors Revisited: Submission to the SoccerNet Challenge 2022},
  howpublished  = {arXiv preprint arXiv:2206.07846},
  year          = {2022},
}

@misc{timm,
  author        = {Ross Wightman and Hugo Touvron and Hervé Jégou},
  title         = {{ResNet} strikes back: An improved training procedure in timm},
  howpublished  = {arXiv preprint arXiv:2110.00476},
  year          = {2021},
}

@misc{baidu,
  author        = {Xin Zhou and Le Kang and Zhiyu Cheng and Bo He and Jingyu Xin},
  title         = {Feature Combination Meets Attention: Baidu Soccer Embeddings and Transformer based Temporal Detection},
  howpublished  = {arXiv preprint arXiv:2106.14447},
  year          = {2021},
}

@misc{gru,
  author        = {Chung, Junyoung and Gulcehre, Caglar and Cho, KyungHyun and Bengio, Yoshua},
  title         = {Empirical {{Evaluation}} of {{Gated Recurrent Neural Networks}} on {{Sequence Modeling}}},
  howpublished  = {arXiv preprint arXiv:1412.3555},
  year          = {2014},
}

@misc{vjepa2,
  author        = {Assran, Mahmoud and Bardes, Adrien and Fan, David and Garrido, Quentin and Howes, Russell and
Komeili, Mojtaba and Muckley, Matthew and Rizvi, Ammar and Roberts, Claire and Sinha, Koustuv and Zholus, Artem and
Arnaud, Sergio and Gejji, Abha and Martin, Ada and Robert Hogan, Francois and Dugas, Daniel and
Bojanowski, Piotr and Khalidov, Vasil and Labatut, Patrick and Massa, Francisco and Szafraniec, Marc and
Krishnakumar, Kapil and Li, Yong and Ma, Xiaodong and Chandar, Sarath and Meier, Franziska and LeCun, Yann and
Rabbat, Michael and Ballas, Nicolas},
  title         = {{V-JEPA~2}: Self-Supervised Video Models Enable Understanding, Prediction and Planning},
  howpublished  = {arXiv preprint arXiv:2506.09985},
  year          = {2025}
}

@misc{vjepa2_1,
  author        = {Mur-Labadia, Lorenzo and Muckley, Matthew and Bar, Amir and Assran, Mahmoud and
Sinha, Koustuv and Rabbat, Michael and LeCun, Yann and Ballas, Nicolas and Bardes, Adrien},
  title         = {V-JEPA 2.1: Unlocking Dense Features in Video Self-Supervised Learning},
  howpublished  = {arXiv preprint arXiv:2603.14482},
  year          = {2026}
}

@misc{hoi_detr,
  author        = {Darkhalil, Ahmad and Damen, Dima and Fouhey, David},
  title         = {Improving and Evaluating Hand-Object Interaction Detection},
  howpublished  = {arXiv preprint arXiv:2606.17384},
  year          = {2026}
}

@misc{qwen25vl,
  author        = {Shuai Bai and Keqin Chen and Xuejing Liu and Jialin Wang and Wenbin Ge and Sibo Song and Kai Dang and Peng Wang and Shijie Wang and Jun Tang and Humen Zhong and Yuanzhi Zhu and Mingkun Yang and Zhaohai Li and Jianqiang Wan and Pengfei Wang and Wei Ding and Zheren Fu and Yiheng Xu and Jiabo Ye and Xi Zhang and Tianbao Xie and Zesen Cheng and Hang Zhang and Zhibo Yang and Haiyang Xu and Junyang Lin},
  title         = {{Qwen2.5-VL} Technical Report},
  howpublished  = {arXiv preprint arXiv:2502.13923},
  year          = {2025},
}

@string{cvpr      = "Proceedings of the {IEEE} Conference on Computer Vision and Pattern Recognition"}

@string{cvprw     = "Proceedings of the {IEEE} Conference on Computer Vision and Pattern Recognition Workshops"}

@string{cvprf     = "Proceedings of the {IEEE} Conference on Computer Vision and Pattern Recognition Findings"}

@string{eccv      = "Proceedings of the European Conference on Computer Vision"}

@string{iccv      =  "Proceedings of the International Conference on Computer Vision"}

@string{iccvw     =  "Proceedings of the International Conference on Computer Vision Workshops"}

@string{nips      = "Advances in Neural Information Processing Systems"}

@string{bmvc      = "Proceedings of the British Machine Vision Conference"}

@string{wacv      = "Proceedings of the IEEE Winter Conference on Applications of Computer Vision"}

@string{wacvw     = "Proceedings of the IEEE Winter Conference on Applications of Computer Vision Workshops"}

@string{ijcnn     = "Proceedings of the International Joint Conference on Neural Networks"}

@string{iclr      = "Proceedings of International Conference on Learning and Representation"}

@string{ijcv      = "International Journal of Computer Vision"}

@string{tpami     = "IEEE Transactions on Pattern Analysis and Machine Intelligence"}

@string{it        = "IEEE Information Theory"}

@string{tsp       = "IEEE Transactions on Signal Processing"}

@string{jun = "June"}
\end{document}